\documentclass[10pt,journal,compsoc,twoside]{IEEEtran}
\usepackage{enumitem}
\usepackage[flushleft]{threeparttable}
\ifCLASSOPTIONcompsoc
\usepackage[nocompress]{cite}
\else
\usepackage{cite}
\fi
\ifCLASSINFOpdf
\usepackage{graphicx}
\usepackage{subcaption}
\graphicspath{ {./figs/} }
\else
\fi
\usepackage{amsmath}
\usepackage{algorithm}
\usepackage{algpseudocode}
\newcommand\Algphase[1]{%
	\vspace*{-.7\baselineskip}\Statex\hspace*{\dimexpr-\algorithmicindent-2pt\relax}\rule{0.5\textwidth}{0.4pt}%
	\Statex\hspace*{-\algorithmicindent}\textbf{#1}%
	\vspace*{-.7\baselineskip}\Statex\hspace*{\dimexpr-\algorithmicindent-2pt\relax}\rule{0.5\textwidth}{0.4pt}%
}
\algnewcommand\algorithmicinput{\textbf{Input:}}
\algnewcommand\Input{\item[\algorithmicinput]}
\algnewcommand\algorithmicoutput{\textbf{Output:}}
\algnewcommand\Output{\item[\algorithmicoutput]}

\usepackage{url}

\usepackage{array, multirow, multicol}
\begin{document}
%
\title{Bayesian Joint Matrix Decomposition for Data Integration with Heterogeneous Noise}

%
%
%
%

\author{Chihao~Zhang
	and~Shihua~Zhang
	\IEEEcompsocitemizethanks{\IEEEcompsocthanksitem Chihua Zhang and Shihua Zhang* are with the NCMIS, CEMS, RCSDS, Academy of Mathematics and Systems Science, Chinese Academy of Sciences, Beijing 100190, China, and School of Mathematics Sciences, University of Chinese Academy of Sciences, Beijing 100049, China. \protect\\ *To whom correspondence should be addressed. Email: zsh@amss.ac.cn.}}


%
%

\markboth{ZHANG C, ZHANG S.:  BAYESIAN JOINT MATRIX DECOMPOSITION FOR DATA INTEGRATION WITH HETEROGENEOUS NOISE}%
{IEEE TRANSACTIONS ON JOURNAL NAME,  MANUSCRIPT ID}
%



\IEEEtitleabstractindextext{%
\begin{abstract}

Matrix decomposition is a popular and fundamental approach in machine learning and data mining. It has been successfully applied into various fields. Most matrix decomposition methods focus on decomposing a data matrix from one single source. However, it is common that data are from different sources with heterogeneous noise. A few of matrix decomposition methods have been extended for such multi-view data integration and pattern discovery. While only few methods were designed to consider the heterogeneity of noise in such multi-view data for data integration explicitly. To this end, we propose a joint matrix decomposition framework (BJMD), which models the heterogeneity of noise by Gaussian distribution in a Bayesian framework. We develop two algorithms to solve this model: one is a variational Bayesian inference algorithm, which makes full use of the posterior distribution; and another is a maximum a posterior algorithm, which is more scalable and can be easily paralleled. Extensive experiments on synthetic and real-world datasets demonstrate that BJMD considering the heterogeneity of noise is superior or competitive to the state-of-the-art methods.
\end{abstract}

}

\maketitle
\IEEEdisplaynontitleabstractindextext
\IEEEpeerreviewmaketitle

\IEEEdisplaynontitleabstractindextext

%
\IEEEpeerreviewmaketitle

\IEEEraisesectionheading{\section{Introduction}\label{sec:introduction}}

%
%
%
%
\IEEEPARstart{M}{atrix} decomposition (a.k.a. matrix factorization) plays fundamental roles in machine learning and data mining. As we know, the basic task of machine learning and data mining is to discover knowledge from primitive data. In many cases, primitive datasets or observations are organized in matrix forms. For example, an image can be stored in a matrix of pixels; a corpus is often represented by a document-term matrix that describes the frequency of terms occur in the collection of documents. The primitive datasets are generally big and noisy matrices. Extracting useful information from primitive datasets directly is often infeasible. Therefore, matrix decomposition is becoming a powerful tool to get a more compact and meaningful representation of observed matrices. Matrix decomposition has been successfully applied to various fields including image analysis \cite{Lee1999, Hoyer2004, liu2012constrained}, gene expression analysis \cite{Brunet2004, carmona2006biclustering, witten2009penalized}, collaborative filtering \cite{srebro2003weighted, hofmann2004latent, srebro2005maximum}, text mining \cite{pauca2004text, ding2006orthogonal} and many other areas.
\par
Generally, most matrix decomposition methods assume a data matrix is from one single data source. The conceptual idea underlying matrix decomposition is that the observed data matrix can be approximated by a product of two or more low-rank matrices. Given an observed data matrix, a general matrix decomposition method can be summarized as follows:
\begin{equation}
 X \approx WH
\end{equation}
which is further formulated into an optimization problem:
\begin{equation}
\min_{W,H}\text{D}(X||WH)
\end{equation}
where $\text{D}$ is a divergence function. Notice that for each column of $X$, $x_{\cdot j} \approx \sum_{k=1}^{K} w_{\cdot k }h_{kj}$, where $K$ is the number of columns of $W$, i.e. each column of the observed matrix is approximated by a linear combination of all columns of $W$. Therefore,  $W$ is often referred as a basis matrix, while $H$ is often referred as a coefficient matrix.

\par Generally, one can get different matrix decomposition methods by placing different constraints on $W$ and $H$. For example, $K$-means clustering may be considered as the simplest matrix decomposition method \cite{Ding2010}. In $K$-means clustering, $W$ contains clustering centroid matrix, which has no restriction; $H$ is a binary matrix and each column of $H$ indicates the cluster membership by a single one. Semi-nonnegative matrix factorization (semi-NMF) \cite{Ding2010} relaxes the restriction of $K$-means. It only restricts $H$ to be nonnegative. The nonnegative restriction on $H$ usually improves the interpretability of the basis matrix. To decompose nonnegative data like images, nonnegative matrix factorization (NMF) has been explored \cite{Paatero1994,Lee1999}. NMF restricts both $W$ and $H$ to be nonnegative. The nonnegative constraints lead to part-based feature extraction and sort of sparseness naturally \cite{Lee1999}. Moreover, due to its effectiveness in clustering and data representation, NMF has been extended to many variant forms including various sparse NMF models \cite{Hoyer2004, Kim2007} and regularized NMF models \cite{Cai2008NonnegativeMF, Cai2011}.

\par However, there are several main drawbacks of those standard matrix factorization methods. First, they are prone to overfitting the observed data matrices. Second, the regularization parameters of models require to be carefully tuned. Bayesian approaches were imposed to matrix factorization \cite{Bishop1999bayesian, welling2006bayesian, schmidt2009bayesian, Salakhutdinova, Saddiki2015} to address those weaknesses by incorporating priors with model parameters and hyperparameters. Third, the above methods only focus on a data matrix from only one source.

\par Currently, multi-view data collected from different sources is very common. Such data from different sources might be complementary to each other. Moreover, they are usually of different quality (i.e. noise level is different in each source). How to make full use of those data is an important but challenging topic. Many studies have devoted their efforts to developing mathematical methods to tackle the underlying issues for data integration. For example, Zhang \textit{et al}. \cite{Zhang2012} proposed joint NMF (jNMF) to perform integrative analysis of multiple types of genomic data to discover the combinatorial patterns that would have been ignored with only a single type of data. jNMF assumes the different data shares a common basis matrix for each data set. Moreover, Zhang \textit{et al}. \cite{Zhang2011} also developed a sparsity-constrained and network-regularized joint NMF for multiple genomics data integration. Liu \textit{et al}. \cite{Liua} developed multi-view NMF (MultiNMF) for multi-view clustering. Unlike jNMF and others shared one-side factor among multi-view data matrices, MultiNMF uses a consensus coefficient matrix to give a multi-view clustering results. More recently, Chalise and Fridley \cite{Chalise2017} proposed a weighted joint NMF model (intNMF) for integrative analysis, which assigns higher weight to data from more reliable source. However, few of these methods address the heterogeneity of noise explicitly. jNMF ignores the noise heterogeneity, MultiNMF and intNMF address this problem implicitly by giving a lower weight to more noisy data source, but the weight of each source is difficult to decide. Thus, powerful computational methods to well consider the heterogeneous noise of multi-view data for data integration are urgently needed.

\par To this end, we propose a powerful joint matrix decomposition framework (BJMD), which models the heterogeneity of noise explicitly by Gaussian distribution in a Bayesian framework. Mathematically, it still remains the interpretability and sparsity of the basis matrix as regularized matrix decomposition does. We develop two algorithms to solve this model: one is a variational Bayesian inference algorithm, which makes full use of the proposal distribution to approximate posterior distribution; and another is a maximum a posterior algorithm (MAP), which turns to be an iterative optimization algorithm, enabling it is more efficient and scalable. We extensively compare the two algorithms and discuss their advantages and disadvantages. Extensive experiments on synthetic and real-world datasets show the effectiveness of BJMD for modeling the noise and demonstrate that considering the heterogeneity of noise leads to superior performance to the state-of-the-art methods.

\section{Related Work}
\subsection{Joint NMF and Other Variants}
\par Joint NMF (jNMF) is a natural extension to NMF for integrating multiple datasets\cite{Zhang2012}. For $C$ data matrices $(X^{(1)})_{M\times N_1},\ldots, (X^{(C)})_{M\times N_C}$, the optimization format of jNMF is formulated as:
\begin{equation}
\begin{aligned}
& \min &  \sum_{c=1}^C ||X^{(c)} - WH^{(c)}||^2_F \\
& \text{s.t.} & W \geq 0, \; H^{(c)} \geq 0  \quad\quad
\end{aligned}
\end{equation}
where $W_{M\times K}$ is the common or shared basis matrix and $H^{(c)}_{K\times N_c}$ is the corresponding coefficient matrix. The common or shared patterns in the multi-view data matrices from different sources can be reflected by the shared basis matrix. One can easily see that jNMF is mathematically equivalent to NMF by setting $\hat{X}=(X^{(1)}, \ldots, X^{(C)})$ and $\hat{H}=(H^{(1)},\ldots, H^{(C)})$. Therefore, jNMF ignores the heterogeneity of noise in different data sources. Stra\u{z}ar \textit{et al}. \cite{Strazar2016} extended jNMF by adding orthogonality-regularized terms on coefficient matrices to predict protein-RNA interactions (iONMF). The orthogonality regularization prevents multicollinearity and iONMF was stated to outperform other NMF models in predicting protein-RNA interactions. However, the heterogeneity of noise among different data types is still ignored.

\par
MultiNMF extends jNMF to multi-view clustering and requires the coefficient matrices learned from various views to be approximately common. Specifically, it is formulated as follows:
\begin{equation}\label{multi_nmf}
\begin{aligned}
& \min & \sum_{c=1}^C ||X^{(c)} - W^{(c)}H^{(c)}||^2_F + \sum_{c=1}^C \lambda_c ||H^{(c)} - H^*||_F^2 \\
& \text{s.t.} &  ||w_{\cdot k}^{(c)}||_1 = 1 \forall k, \text{ and }  W^{(c)} \geq 0, H^{(c)} \geq 0, H^* \geq 0
\end{aligned}
\end{equation}
where $H^*$ is the consensus coefficient matrix and $\lambda_c$  is the  weight parameter to tune the relative importance among different views. In image processing, Jing \textit{et al}. \cite{jing2012snmfca} proposed a supervised joint matrix factorization model for image classification and annotation (SNMFCA). SNMFCA factorizes region-image matrix and annotation-image matrix simultaneously and incorporates the label information as a network-regularized term. SNMFCA also uses weight hyperparameters to balance the importance of each data source. Recently, Chalise and Fridley \cite{Chalise2017} proposed intNMF which is an extension of jNMF by specifying weight hyperparameter to each data source. Specifically, it is formulated as follows:
\begin{equation}\label{intNMF}
\begin{aligned}
& \min & \sum_{c=1}^C \lambda_c ||X^{(c)} - WH^{(c)}||^2_F  \\
& \text{s.t.} &    W \geq 0, H^{(c)} \geq 0  \quad\quad
\end{aligned}
\end{equation}
where $\lambda_c$ is a user predefined weight parameter. Those methods use weight parameters to regularize the importance of each data source. However, how to choose those weight parameters is still not sufficiently studied in literature.

\subsection{GLAD}
\par Bayesian methods for matrix decomposition have been studied for about twenty years. In 1999, Bishop \cite{Bishop1999bayesian} proposed Bayesian principal component analysis (BPCA), which can automatically determine an effective number of principal components as part of the inference procedure. Later, Bishop \cite{bishop1999variational} gives an effective variational inference method for BPCA. Welling and Kurihara \cite{welling2006bayesian} introduced Bayesian $K$-means, which retains the ability of selecting the model structure by incorporating prior on model parameters. Schmidt \textit{et al}. \cite{schmidt2009bayesian} proposed a Bayesian variant for NMF by introducing exponential priors on the basis matrix and coefficient matrix respectively (BNMF). Moreover, BNMF assumes Gaussian noise over the product of the basis matrix and coefficient matrix. However, the Gaussian noise may lead to the entry of data matrix to be negative. In collaborative filtering, Salakhutdinov and Mnih \cite{Salakhutdinova} proposed Bayesian probabilistic matrix factorization (BPMF), which places Gaussian prior over both basis and coefficient matrices to predict user preference for movies. BPMF has attracted great interests of researchers from diverse fields. Moreover, other Bayesian methods for matrix decomposition in collaborative filtering field have also been proposed \cite{Lawrence2009, porteous2010bayesian, mackey2010mixed}.

\par GLAD \cite{Saddiki2015} is a mixed-membership model, which utilizes three typical distributions (i.e., Gaussian distribution, Laplace distribution and Dirichlet distribution). GLAD can be described as a matrix decomposition model with Laplace prior on the basis matrix, Dirichlet prior on the coefficient matrix and Gaussian noise on the observed data matrix. Specifically, GLAD assumes that the observed data matrix $X_{M\times N}$ is generated as follows:
\begin{equation}
X = WH + \epsilon
\end{equation}
where each element of the basis matrix $w_{ik} \sim \text{L}(0, \lambda)$, each column of the coefficient matrix $h_{\cdot j} \sim \text{Dir}(\alpha)$ and $\epsilon$ indicates the Gaussian noise $\epsilon \sim \text{N}(0, \sigma^2)$. Mathematically, the Laplace prior over $w_{ik}$ enforces the sparsity of the basis matrix. And $h_{\cdot j}$ has a Dirichlet distribution, providing a distribution over $K$ clusters for each column of $X$. Inference for GLAD focuses on estimating the posterior distribution over latent variables:
\begin{equation}
p(W,H|X; \alpha, \lambda, \sigma^2)=\frac{p(W,H, X; \alpha, \lambda, \sigma^2)}{p(X; \alpha, \lambda, \sigma^2)}
\end{equation}
\par However, due to the non-conjugate prior, it is intractable to calculate the posterior distribution, which involves high dimensional integral. GLAD seeks to a non-conjugate variational inference (VI). Instead of calculating the integral exactly, VI uses a distribution $q(W,H;\phi)$ parameterized by $\phi$, to approximate the true posterior distribution. Specifically, VI aims at minimizing the Kullback-Leibler (KL) divergence between the approximation distribution  $q(W,H)$ and the true posterior distribution $p(W,H|X; \alpha, \lambda, \sigma^2)$:
\begin{equation}
\underset{q \in P}{\min \textbf{KL}(q||p)}= -\int q(W,H) \ln{\frac{p(W,H|X; \alpha, \lambda, \sigma^2)}{q(W,H;\phi)}}dWdH
\end{equation}
where $\textbf{KL}(q||p)$ denotes the KL divergence between $q(W,H;\phi)$ and $p(W,H|X; \alpha, \lambda, \sigma^2)$, and $P$ denotes the family of probability distribution functions (PDFs) to make the integral trackable. Because of involving posterior distribution, the KL divergence is still difficult to optimize. Instead, VI maximize the evidence lower bound (ELBO):
\begin{equation}
\label{elbo}
\mathcal{L} = \text{E}_q[\ln p(X,W,H)] - \text{E}_q[\ln p(W,H)]
\end{equation}
where the first term is the expectation of joint distribution over the approximation distribution, and the second term is the entropy of approximation distribution. Therefore, if the expectation of (\ref{elbo}) can be calculated analytically, VI turns the inference problem to be an optimization problem. Unfortunately, expectation is still difficult to compute for GLAD. To address this issue, GLAD further introduces a Laplace approximation to address the expectation in ELBO.

\par The inference of GLAD has two main drawbacks. First, it approximates the ELBO, which is only an approximation of true posterior. Consequently, the gap between approximation distribution and true posterior distribution might be widened. Second, it is still computationally expensive for involving a non-linear optimization problem at each iteration. Thus, GLAD takes considerable time to converge for even small data, which makes it unrealistic for real-world datasets.

\section{Bayesian Joint Matrix Decomposition}
\par
\begin{figure}[!t]

	\centering
	\includegraphics[width=\columnwidth]{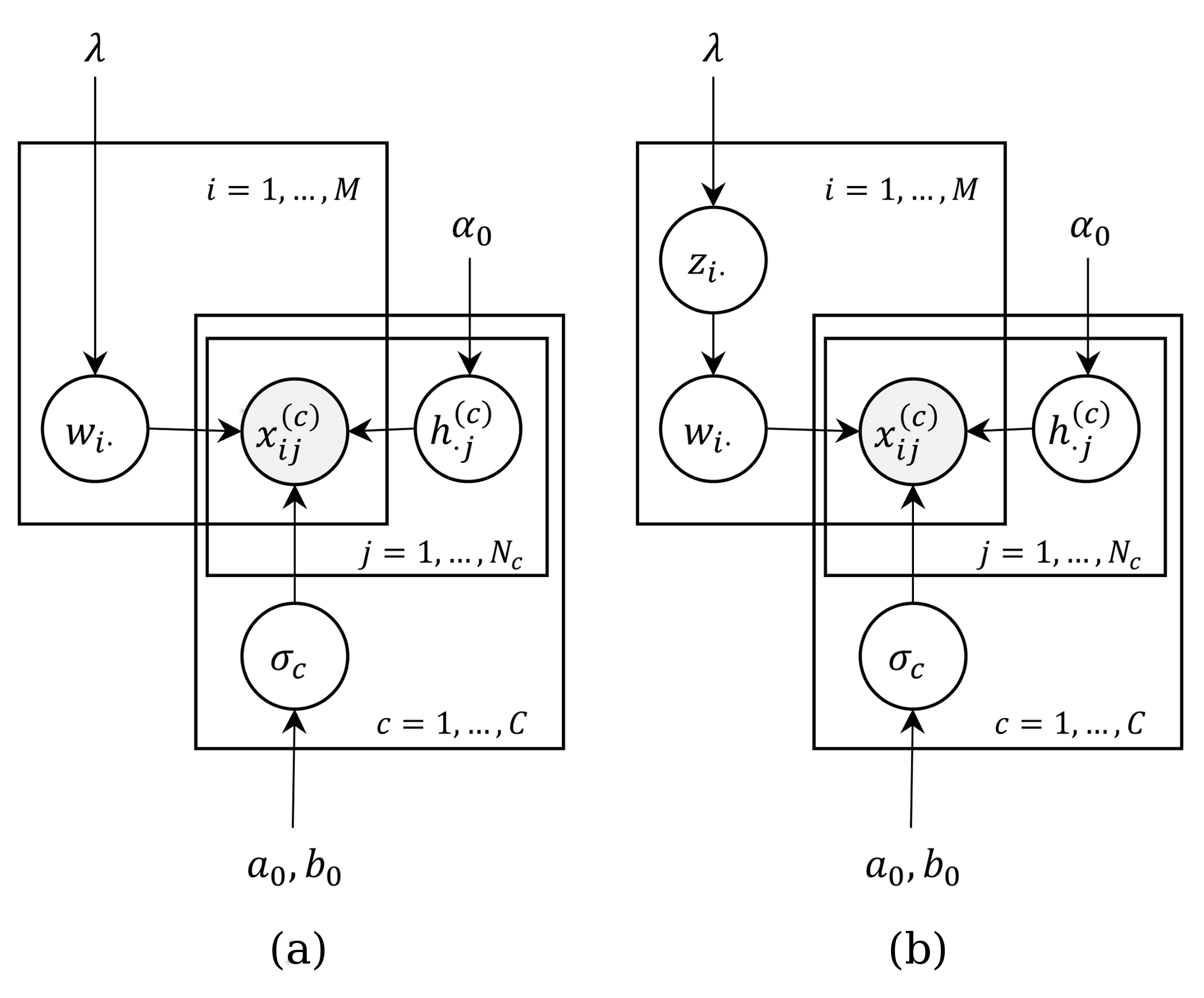}
	\caption{Graphical models of (a) BJMD and (b) reformulated BJMD.}
	\label{fig_grahical}
\end{figure}

 In this section, we present the Bayesian joint matrix decomposition framework (BJMD) (Fig. \ref{fig_grahical}a), and develop two algorithms to solve it.

\subsection{Model Construction}

\par Suppose that we have multi-view data matrices: $(X^{(1)})_{M\times N_1},\ldots,(X^{(C)})_{M\times N_C}$. Inspired by previous studies, the generative model of BJMD is formulated as follows:
\begin{equation}
X^{(c)} = WH^{(c)} + \epsilon_c \text{,  for $c=1,\ldots, C$}
\end{equation}
where  $W$ is a shared basis matrix and $\epsilon_c$ models noise of different data sources by Gaussian distribution. To enforce the sparsity of the basis matrix\cite{Kaban2007}, we place a Laplace prior with zero mean on it:
\begin{equation}
w_{ik} \sim p(w_{ik}|0, \lambda)
\end{equation}
where
\begin{equation}
p(y|\mu, \lambda)=\frac{1}{2\lambda}\exp \left(-\frac{|y-\mu|}{\lambda}\right)
\end{equation}

\par Non-negative restriction on the coefficient matrix of semi-NMF \cite{Ding2010} leads to better interpretability of the basis matrix. However, columns of the basis matrix are usually not comparable without proper normalization, which weakens the interpretability. In fact, one can see that semi-NMF is invariant with column scaling on  $W$ or row scaling on $H$. Consider a diagonal matrix $S$ with size $K \times K$, $WH$ is equal to $WSS^{-1}H$. To this end, we additionally restrict each column of coefficient matrices summing to one. As the support of Dirichlet distribution is a unit simplex, we place a Dirichlet prior on each column of coefficient matrices $H^{(c)}$:
\begin{equation}
h_{\cdot j}^{(c)} \sim p(h_{\cdot j}^{(c)}|\alpha_0)
\end{equation}
where
\begin{equation}
p(y|\alpha_0) = \frac{\Gamma(\sum_{i=1}^K) \alpha_{0i}}{\prod_{i=1}^K\Gamma(\alpha_{0i})}\prod_{i=1}^K y_i^{\alpha_{0i} - 1}
\end{equation}
 $\alpha_{0}$ is a $K$-dimensional vector and  $\alpha_{0i}$ indicates the  $i$th element of $\alpha_{0}$. Moreover, we use Gaussian distribution in each data source modeling the heterogeneity of noise explicitly:
\begin{equation}
\epsilon_c \sim p(\epsilon_c|0, \sigma_c^2)
\end{equation}
where
\begin{equation}
p(y|\mu, \sigma_c^2) = \frac{1}{\sqrt{2\pi}\sigma_c}\exp \left(-\frac{(y-\mu)^2}{2\sigma_c^2} \right)
\end{equation}
To get a complete Bayesian model, we also need to introduce prior distribution for variance of Gaussian distribution. For convenience, we choose conjugate prior for $\sigma_c^2$, i.e. $\sigma_c^2 \sim IG(a_0, b_0)$, where $a_0$, $b_0$  are two hyperparameters of inverse Gamma distribution and can be set to a small positive value such as $1$ or $0.01$ in a noninformative fashion \cite{Gelman2006}.

\par Under the BJMD model, the posterior is proportional to the complete likelihood, which can be written as:
\begin{flalign}
\begin{split}
& p(W, H^{(1)}\ldots H^{(C)}, \sigma_1^2 \ldots \sigma_C^2, X^{(1)} \ldots X^{(C)};\lambda, \alpha_0, a_0, b_0)  \\
= & p(W;\lambda)\prod_{c=1}^C p(X^{(c)}|W, H^{(c)}, \sigma_c^2)p(H^{(c)};\alpha_0)p(\sigma_c^2;a_0, b_0)\\
= & \prod_{i,k} p(w_{ik};\lambda)\prod_{c=1}^C \left( \prod_{j=1}^{N_c} p(h^{(c)}_{\cdot j};\alpha_0)\right) \\
& \left(\prod_{i,j}p(x_{ij}|w_{i\cdot}h^{(c)}_{\cdot j}, \sigma_c^2)p(\sigma_c^2;a_0,b_0)\right)
\end{split}
\end{flalign}
which could help us to understand it from an optimization perspective. Hence, we write down the log-likelihood. For simplicity, we formulate the complete negative log-likelihood retaining only terms involving $W$, $H^{(c)}$  as follows:
\begin{flalign}\label{opt1}
\begin{split}
& -LL(W,H^{(1)}, \ldots, H^{(C)}|X^{(1)}, \ldots, X^{(C)}) \\
= & \sum_{c=1}^{C}\sum_{i,j}\frac{1}{2\sigma_c^2}(x_{ij}^{(c)} - w_{i\cdot}h^{(c)}_{\cdot j})^2 + \sum_{i,k}\frac{|w_{ik}|}{\lambda} \\
 & - \sum_{c=1}^{C}\sum_{k,j}(\alpha_{0k} - 1)\ln h^{(c)}_{kj}
 \end{split}
\end{flalign}
with the unit simplex constraints $\sum_{k=1}^K h_{kj}^{(c)}=1 $ and $ h_{kj}^{(c)} \geq 0$ for any $k$, $j$, $c$. The first term in (\ref{opt1}) can be viewed as a divergence function between the observation $x_{ij}^{(c)}$ and the approximated value $w_{i\cdot}h^{(c)}_{\cdot j}$. The variance of Gaussian distribution $\sigma_c^2$ is the weight parameter, which gives a higher weight for the data points from source with lower noise level. Different from the weight hyperparameters of intNMF and MultiNMF,  $\sigma_c^2$ can be automatically inferred during the optimization process. The second term of this equation is a $L_1$-norm regularization on the basis matrix $W$, which regularizes the sparsity of $W$. The last term is a regularization term on the coefficient matrices. If we omit the first two terms, one can find that the last term is minimized when $h^{(c)}_{kj} = \alpha_{0k} / \sum_{k=1}^K \alpha_{0k}$, which is actually the expectation of Dirichlet prior. Therefore, the last term enforces $h_{kj}^{(c)}$  to the expectation of prior and reduces the risk of overfitting.

\subsection{Variational Inference}
\par In this subsection, we adopt a variational inference algorithm for the BJMD model. For further comparison, we first extend the GLAD to multi-view data matrices by a shared basis matrix and different coefficient matrices. Note that the GLAD and the extended GLAD model for multi-view data are not in a full Bayesian framework. Both models treat the variance of Gaussian distribution as a hyperparameter rather than a distribution. Thus, we refer the extended GLAD model as joint matrix decomposition (JMD) later.
\par We can easily see that the inference algorithm of GLAD can be naturally extended to JMD. However, the extended inference algorithm of JMD still suffers the aforementioned problems of GLAD: (a) the approximation of ELBO might not be a good approximation to the true posterior; (b) the expensive computational cost makes it unrealistic for real-world datasets. Recent advances in stochastic variational methods, including black box variational inference (BBVI) \cite{Ranganath}, auto-encoding variational Bayes (AEVB) [18] and automatic differentiation variational inference (ADVI) \cite{Kingma2013}, provide powerful tools to tackle those problems. These stochastic variational inference methods avoid calculating the ELBO directly. Instead, they take derivative of ELBO and push the derivative operator into the expectation:
\begin{equation}\label{gELBO}
\nabla \mathcal{L}(\phi) = \text{E}_q[ \nabla_{\phi} \ln q(z;\phi)( \ln p(x,z) - \ln q(z;\phi))]
\end{equation}
where $x$ denotes the observed data, $z$ denotes the latent variable,   $q(z;\phi)$ denotes the variational distribution parameterized by $\phi$, $p(x,z)$ denotes the joint distribution of latent variables and observed data, and $ \mathcal{L}(\phi)$  denotes the ELBO. The gradient of ELBO (\ref{gELBO}) is expressed as an expectation over variational distribution, which can be approximated by a Monte Carlo method:
\begin{equation}
\label{eq_mc}
\nabla \mathcal{L}(\phi) = \frac{1}{S} \sum_{s=1}^S \nabla_{\phi} \ln q(z_s;\phi)( \ln p(x,z_s) - \ln q(z_s;\phi))
\end{equation}
where $S$ is the number of samples and $z_s \sim q(z;\phi)$.  Now we can update the variational parameters with the noisy gradients with simple computation.

\par We implement ADVI for optimizing the BJMD model with python library PyMC3 \cite{salvatier2016probabilistic}. ADVI converts all constrained latent variables to unconstrained ones and uses Gaussian distributions to approximate the ELBO of unconstrained random variables. Following previous notations, ADVI can be briefly summarized as follows:
\begin{enumerate}[label=(\alph*)]
\item Firstly, ADVI transforms the constrained latent variables into unconstrained real-valued variables $\xi$ by a one-to-one mapping. ADVI aims at minimizing $\mathbf{KL}(q(\xi)||p(\xi|x))$ in which one can use a single family of distributions for all models.
\item Then ADVI converts the gradient of variational objective function as expectation over $q(\xi)$, which can be estimated by a Monte Carlo sampling strategy as in (\ref{eq_mc}).
\item Furthermore, ADVI reparameterizes the gradient in terms of standard Gaussian. This transformation turns the Monte Carlo procedure to sampling from standard Gaussian distribution, enabling it can be sampled effectively.
\item Finally, ADVI uses the noisy gradients to optimize the variational distribution. As all transformation mentioned in the former three steps are one-to-one mappings, to get the variational parameters $\phi$ we concern, we can map the variational distribution to the original space of the constrained latent variable $z$.
\end{enumerate}

\par ADVI can automatically infer probabilistic models by the above steps. One should notice that the variational distribution is optimized by noisy gradients, and thus it is possible for ADVI to escape from the local optimum and will not be overfitted. The disadvantage of noisy gradients is that the ELBO will not converge. Instead, the ELBO will fluctuate in a small interval. Therefore, it is difficult to decide when to stop iteration. Empirically, stopping iteration immediately after the ELBO becoming stable often leads to poor performance. As we use the coefficient matrices for further clustering and representation, we stop iteration if the coefficient matrices are stable. ADVI empirically takes tens of thousands of iterations to get a well-trained model. Then checking the relative change of basis matrices at each iteration leads to an expensive computational cost. Thus, we check the coefficient matrices every $L$  iterations to ensure that the model is sufficiently trained:
\begin{equation}
\label{stop_advi}
\max_c \frac{||(H^{(c)})^{(T-1)L}||_F^2}{||(H^{(c)})^{TL} - (H^{(c)})^{(T-1)L}||_F^2} < \tau_1
\end{equation}
where $\tau_1$ is a tolerance threshold, which is a small positive value,  $(H^{(c)})^{TL}$ is the coefficient matrices at  $TL$th iteration ($T=1,2,3 \ldots$). Note that relative change of $H^{(c)}$ will not converge to zero due to the noisy gradient. Thus, the iteration should be stopped if the criterion (\ref{stop_advi}) is satisfied or iteration exceeds $iter_{max}$, where  $iter_{max}$ is a user predefined maximum times of iteration.

\subsection{Maximum A Posterior}
\par ADVI makes it possible to apply BJMD framework to real-world datasets, yet it still takes tens of thousands of iterations before stopping criterion is satisfied. To get a more efficient and scalable algorithm, we develop a maximum a posterior (MAP) algorithm, which directly maximizes the posterior in (\ref{opt1}) and treat it as an optimization problem. The key difference between VI and MAP is that: VI approximates the posterior distribution by a family of distribution; MAP approximates posterior distribution by its mode. VI makes full use of posterior distribution, while it pays a more expensive computational cost. In contrast, MAP is a point estimation with less computational cost. We present a MAP algorithm for the BJMD model in the following.

\subsubsection{Model Reformulation}
Due to the absolute value factor in Laplace distribution, it is inconvenient for posterior inference within a Bayesian framework for (\ref{opt1}). To address this computational issue, we reformulate the model by utilizing a two-level hierarchical representation of Laplace distribution. Specifically, a random variable $x$  following a Laplace distribution $\text{L}(x|\mu,\sqrt{\lambda / 2})$ can be reformulated as Gaussian scale mixtures with exponential distribution prior to the variance \cite{Andrews1974}. That is:
\begin{equation}
\begin{split}
p(x|\mu, \sqrt{\lambda /2}) & = \frac{1}{2}\sqrt{\frac{2}{\lambda}} \exp \left(-\frac{2}{\lambda}|x-\mu|\right) \\
& = \int_0^{\infty} \frac{1}{\sqrt{2\pi} z} \exp \left(-\frac{(x-\mu)^2}{2z}\right)\frac{1}{\lambda} \exp \left( -\frac{z}{\lambda}\right) dz\\
& = \int_0^{\infty} \text{N}(x|\mu, z) p(z|\lambda) dz
\end{split}
\end{equation}
where the term $ p(z|\lambda)= 1/\lambda \exp (-z /\lambda) $ is the PDF of exponential distribution.
\par Therefore, we can replace the Laplace prior of $W$  with the two-level hierarchical representation to eliminate the absolute value factors. This is a useful technique to simplify computation, which has been used in other studies \cite{zhao2015l}. Under this reformulation, the complete log likelihood is:
\begin{flalign}
\begin{split}
\label{eq_rp}
& p(W,Z,H^{(1)}\ldots H^{(C)}, \sigma_1^2 \ldots \sigma_C^2, X^{(1)} \ldots X^{(C)};\lambda, \alpha_0, a_0, b_0)  \\
= & \prod_{i,k} p(w_{ik}|z_{ik})p(z_{ik};\lambda)\prod_{c=1}^C \left( \prod_{j=1}^{N_c} p(h^{(c)}_{\cdot j};\alpha_0)\right) \\
& \left(\prod_{i,j}p(x_{ij}|w_{i\cdot}h^{(c)}_{\cdot j}, \sigma_c^2)p(\sigma_c^2;a_0,b_0)\right)
\end{split}
\end{flalign}
where random variable $z_{ik}$ is an auxiliary variable. Substituting corresponding PDF into (\ref{eq_rp}), the negative log complete likelihood is formulated as follows:
\begin{equation}
\label{eq_rll}
\begin{split}
& -LL(W,Z,H^{(1)}\ldots H^{(C)}, \sigma_1^2 \ldots \sigma_C^2, X^{(1)} \ldots X^{(C)};\lambda, \alpha_0, a_0, b_0)  \\
& = \sum_{c=1}^C \sum_{i,j} \frac{1}{2\sigma_c^2}(x_{ij}^{(c)} -w_{i\cdot}h^{(c)}_{\cdot j})^2 - \sum_{c=1}^C \sum_{k,j} (\alpha_{0k} -1)\ln h^{(c)}_{kj}  \\
& + \sum_{i,k} \frac{z_{ik}}{\lambda} + \frac{1}{2} \sum_{i,k} \ln z_{ik} + \sum_{i,k} \frac{w_{ik}^2}{2z_{ik}} - \sum_{c=1}^C \ln p(\sigma_c^2;a_0,b_0)
\end{split}
\end{equation}
with the unit simplicial constraints $\sum_{k=1}^K h^{(c)}_{kj} = 1$ and $ h^{(c)}_{kj} \geq 0$ for any $k$, $j$, $c$. Note that the absolute value bars are eliminated after reformulation. Instead, the original $L_1$-norm regularized term is converted to a weighted $L_2$-norm regularized term, which is easier to optimize. Fig. \ref{fig_grahical}b shows the graphical model of the reformulated BJMD.

\par Equivalently, we minimize the complete non-negative log-likelihood (\ref{eq_rll}) in the following optimization form:
\begin{equation}
\label{opt2}
\begin{split}
\min & F= \sum_{c=1}^C \sum_{i,j} \frac{1}{2\sigma_c^2}(x_{ij}^{(c)} -w_{i\cdot}h^{(c)}_{\cdot j})^2  \\
&  - \sum_{c=1}^C \sum_{k,j} (\alpha_{0k} -1)\ln h^{(c)}_{kj}  + \sum_{i,k} \frac{z_{ik}}{\lambda}  \\
& + \frac{1}{2} \sum_{i,k} \ln z_{ik} + \sum_{i,k} \frac{w_{ik}^2}{2z_{ik}} - \sum_{c=1}^C \ln p(\sigma_c^2;a_0,b_0) \\
\text{s.t. } & Z > 0, H^{(c)} > 0  \\
& \sum_{k=1}^K h_{kj}^{(c)} = 1
\end{split}
\end{equation}

\subsubsection{Update Rules}
\par Here, we propose a powerful iterative update procedure to solve problem (\ref{opt2}), which can be easily paralleled. Note that this optimization problem  (\ref{opt2}) is not convex. Therefore, we optimize it in a block-coordinate manner. Specifically, we update one latent variable with others fixed until convergence.\\
\textbf{Updating $Z$ and fixing others:} Let the partial derivative for $z_{ik}$ be zeros, $\frac{\partial F}{\partial z_{ik}} = \frac{1}{\lambda} + \frac{1}{2z_{ik}} - \frac{w_{ik}^2}{2z_{ik}^2}=0$. This is a simple quadratic equation in one variable. Considering the constraint $z_{ik} > 0$, we have:
\begin{equation}
\label{update_z}
z_{ik} = \frac{\sqrt{\lambda^2 + 8w_{ik}^2\lambda} -  \lambda}{4}
\end{equation}
\\
\textbf{Updating $W$ and fixing others:} We only care about $W$ at this step. Therefore, retaining terms only involving $W$, the optimization problem is equivalent to:
\begin{equation}
\min_W \sum_{c=1}^C \sum_{i,j} \frac{1}{2\sigma_c^2}(x_{ij}^{(c)} -w_{i\cdot}h^{(c)}_{\cdot j})^2  + \sum_{i,k} \frac{w_{ik}^2}{2z_{ik}}
\end{equation}
with no constraints. Note that the above optimization problem is separable. For each row $w_{i\cdot}$ of $W$, the subproblem can be written as:
\begin{equation}
\label{opt_w}
\min_{w_{i\cdot}} \sum_{c=1}^C \frac{1}{2\sigma_c^2}||x_{i\cdot}^{(c)} -w_{i\cdot}H^{(c)}||^2_2 + \frac{1}{2}||w_{i\cdot} \text{diag}(\sqrt{z_{i\cdot}})^{-1}||^2_2
\end{equation}
where $\text{diag}(\cdot)$ denotes the diagonal function and $\sqrt{\cdot}$ denotes the element-wise root of square. This subproblem is close to ridge regression \cite{Hoerl1970}. The difference is that the elements of vector $z_{i\cdot}$ are equal in ridge regression. The diagonal matrix $\text{diag}(\sqrt{z_{i\cdot}})^{-1}$ has the special form $\epsilon^2 I$. While in our case, each element of $z_{i\cdot}$ follows the exponential distribution with parameter  $\lambda$. Let the derivative of (\ref{opt_w}) be zero:
\begin{equation}
\sum_{c=1}^C \frac{1}{\sigma_c^2} w_{i\cdot} H^{(c)}H^{(c)T} + w_{i\cdot} \text{diag}(\sqrt{z_{i\cdot}})^{-1} - \sum_{c=1}^C \frac{1}{\sigma_c^2} x_{i\cdot}H^{(c)} = 0
\end{equation}
The optimal solution of subproblem (\ref{opt_w}) is:
\begin{equation}
\label{update_w}
w_{i\cdot} = \Big(\sum_{c=1}^C \frac{1}{\sigma_c^2} x_{i\cdot}H^{(c)}\Big)\Big(\sum_{c=1}^C \frac{1}{\sigma_c^2}H^{(c)}H^{(c)T} + \text{diag}(\sqrt{z_{i\cdot}})^{-1}\Big)^{-1}
\end{equation}
\textbf{Updating $H^{(c)}$ and fixing others:} Each column of $H^{(c)}$ follows a Dirichlet distribution, whose support is a $K$-dimensional simplex. Retaining terms involving $H^{(c)}$, our problem can be written as:
\begin{equation}
\label{opt_h}
\begin{split}
\min_{H^{(c)}} & \sum_{c=1}^C \frac{1}{2\sigma_c^2}||X^{(c)} - WH^{(c)}||^2_F   - \sum_{c=1}^C \sum_{k,j} (\alpha_{0k} -1)\ln h^{(c)}_{kj}\\
\text{s.t. }   & \quad \sum_{k=1}^K h_{kj}^{(c)} = 1 \\
               & \quad h_{kj}^{(c)} > 0
\end{split}
\end{equation}
If $\alpha_0 \geq 1$, the above is a convex optimization problem. Otherwise, it would be difficult to solve this problem due to the non-convexity. We always set $\alpha_0 \geq 1$ in the following experiments. Note that above problem is separable. We can divide it into subproblems by column, and optimize each subproblem in a parallel manner. Subproblem for column $j$ is as follows:
\begin{equation}
\label{sub_opt_h}
\begin{split}
\min_{h^{(c)}_{\cdot j}} & \sum_{c=1}^C \frac{1}{2\sigma_c^2}||Wh^{(c)}_{\cdot j} - x^{(c)}_{\cdot j}||^2_2 - \sum_{k=1}^K (\alpha_{0k} -1)\ln h^{(c)}_{kj}\\
\text{s.t. }  & \quad \sum_{k=1}^K h_{kj}^{(c)} = 1 \\
              & \quad h_{kj}^{(c)} > 0
\end{split}
\end{equation}

\par For convenience, we denote $\alpha_k=\sigma_c^2(\alpha_{0k} - 1)$. The subproblem (\ref{sub_opt_h}) is very close to a quadratic programming problem with unit simplex constraints. The objective function is just a quadratic term coupled with a logarithmic barrier function. The difference is that, for the constraint quadratic programming (QP) problem, $\alpha_k$ is a smooth parameter to enforce the path to be interior within the feasible region. $\alpha_k$  is usually small or gradually decreasing in QP. In our case, $\alpha_k$ is the hyper-parameter related to the Dirichlet prior. If we are confident with our prior knowledge, $\alpha_k$  is not necessarily small.
\par This subproblem can be solved by a interior-point algorithm \cite{Monteiro1989}. Omitting constants, the Lagrange function is:
\begin{equation}
\begin{split}
L(h^{(c)}_{\cdot j}, \mu) = & \frac{1}{2} {h^{(c)}_{\cdot j}}^T(W^TW)h^{(c)}_{\cdot j} - b^Th^{(c)}_{\cdot j} - \sum_{k=1}^K\alpha_k \ln h^{(c)}_{kj} \\
& - \mu(1^T_Kh^{(c)}_{\cdot j}  -1)
\end{split}
\end{equation}
where $b=W^Tx_{\cdot j}^{(c)}$. The KKT condition is as follows:
\begin{align*}
\frac{\partial L}{\partial h^{(c)}_{\cdot j}} & = W^TWh^{(c)}_{\cdot j} -\mu1_K -b -s = 0\\
\frac{\partial L}{\partial \mu} &= 1^T_Kh^{(c)}_{\cdot j}  -1 = 0
\end{align*}
where $s_k =\alpha_k / h^{(c)}_{kj} $. For a given $\alpha_k$, the KKT condition is a system of non-linear equations (the nonlinearity is due to the product $h^{(c)}_{\cdot j} s_k =\alpha_k$). There are $2K+1$ variables $(h^{(c)}_{kj}, \mu, s)$. Let's set
\begin{equation}
F_\alpha(h^{(c)}_{\cdot j}, \mu, s) =
\begin{bmatrix}
W^TW h^{(c)}_{\cdot j} - \mu1_K -b -s \\
1^T_Kh^{(c)}_{\cdot j}  -1 \\
\text{diag}(h^{(c)}_{\cdot j})s - \alpha
\end{bmatrix}
\end{equation}

\par The system  $F_\alpha(h^{(c)}_{\cdot j}, \mu, s)=0$ can be solved in a Newton-like method. At each iteration, to choose a good direction $d$, we use Jacobian matrix as linear approximation. Then, the following  $(2K+1)\times(2K+1)$ linear equation should be solved,
\begin{equation}\label{KKT_system}
J_{F_\alpha}(h^{(c)}_{\cdot j}, \mu, s)d = -F_\alpha(h^{(c)}_{\cdot j}, \mu, s)
\end{equation}
where the Jacobian matrix has the special form:
\begin{equation}
J_{F_\alpha}(h^{(c)}_{\cdot j}, \mu, s) =
\begin{bmatrix}
{X^{(c)}}^T X^{(c)} & -1_K & -I\\
1_K^T & 0 & 0_K^T \\
 \text{diag}(s) & 0_K & \text{diag}(h^{(c)}_{\cdot j})
\end{bmatrix}
\end{equation}
\par
Here we summarize the algorithm of updating  $h^{(c)}_{\cdot j}$ in \textbf{subalgorithm 1}.
\\
\textbf{Updating $\sigma_c^2$ and fixing others:} Due to the conjugate prior, $\sigma_c^2$ can be easily updated by:
\begin{equation}\label{update_sigma}
\sigma_c^2 = \frac{2b_0 +\sum_{i,j} (x^{(c)}_{ij}-w_{i\cdot}h^{(c)}_{\cdot j})^2}{2a_0+MN_c+2}
\end{equation}
\textbf{Stop criterion:} It is easy to see that each step of the algorithm is convex. The value of objective function is ensured to be monotonically non-increasing. We stop iteration when the relative change of objective function is small enough:
\begin{equation}
\label{stop2}
\frac{|f^t-f^{t-1}|}{|f^{t-1}|} \leq \tau_2
\end{equation}
where $f^t$ is the value of objective function at $t$ iteration and $\tau_2$ is the tolerance, which is a small positive value (e.g. $10^{-3}$).
\par
The MAP algorithm updates the latent variables in turn by the above update rules until the stopping criterion (\ref{stop2}) is satisfied. We summarize the MAP algorithm for the BJMD model in \textbf{Algorithm} \ref{algo_map}.
\begin{table}
\label{algo}
\begin{algorithm}[H]
\begin{algorithmic}[1]\caption{\textbf{Maximum a posterior for the BJMD model}}\label{algo_map}
\Input data matrices $X^{(1)}$,$\ldots$, $X^{(C)}$ and initialized $Z$, $W$, $H^{(1)}$, $\ldots$, $H^{(C)}$, $\sigma_1^2$, $\ldots$, $\sigma_C^2$
\Output  $Z$, $W$, $H^{(1)}$, $\ldots$, $H^{(C)}$, $\sigma_1^2$, $\ldots$, $\sigma_C^2$
\Repeat
\For{$i=1,\ldots, M$}\Comment{Can be in  parallel}
\State update $w_{i\cdot}$ by (\ref{update_w})
\EndFor
\For{$c=1,\ldots,C$}
\For{$j=1,\ldots,N_c$}
\Comment{Can be in  parallel}
\State update $h_{\cdot j} ^{(c)}$ by \textbf{Subalgorithm 1}
\EndFor
\EndFor
\State update $Z$ by (\ref{update_z})
\State update $\sigma_1^2$,$\ldots$, $\sigma_C^2$ by (\ref{update_sigma})
\Until{(\ref{stop2}) is satisfied}

\Algphase{Subalgorithm 1 - \text{Interior-point algorithm for} $h_{\cdot j}^{(c)}$}
\Input initialized $(h^{(c)}_{\cdot j}, \mu, s)$, $\eta \in (0,1)$ and other parameters
\Output the $j$th column of coefficient matrix $h^{(c)}_{\cdot j}$
\Repeat
\State solve the system of equations (\ref{KKT_system})
\State determine the searching direction $(\Delta h^{(c)}_{\cdot j}, \Delta  \mu, \Delta  s)$
\State choose the largest step length $\rho$ such that $(h^{(c)}_{\cdot j}, s) + \rho(\Delta h^{(c)}_{\cdot j}, \Delta  s) > 0$
\State set $\rho \gets \min \{1, \eta \rho\}$
\State set $(h^{(c)}_{\cdot j}, \mu, s) \gets (h^{(c)}_{\cdot j}, \mu, s) + \rho(\Delta h^{(c)}_{\cdot j}, \Delta  \mu, \Delta  s)$
\Until{change of objective function (\ref{sub_opt_h}) is small}
\end{algorithmic}
\end{algorithm}
\end{table}

\subsubsection{Computational Complexity}
\par Here we briefly discuss the computational complexity of the proposed MAP algorithm for the BJMD model. At each iteration, the most computational expensive step is inferring $h^{(c)}_{\cdot j}$s, which involves solving a constrained QP problem. To infer $h^{(c)}_{\cdot j}$, each iteration of \textbf{Subalgorithm 1} needs to solve a  $(2K + 1) \times (2K+1)$ linear system, which is of $O((2K+1)^3)$. If QP stops after $q$ iterations, the total cost of updating  $h^{(c)}_{\cdot j}$s is $O(qN(2K+1)^3)$, where $N=\sum_{c=1}^C N_c$. To infer $w_{i\cdot}$, a $K\times K$ linear system should be solved, which leads to $O(MK^3)$ cost in total. The computational cost of updating $Z$ and variance $\sigma_c^2$ are of $O(MK)$ and $O(MNK)$ respectively. Altogether, each iteration of MAP is of $O(qN(2K+1)^3 + MK^3 + MNK)$, and the total complexity of the algorithm is thus $O(T(q  N(2K+1)^3 + MK^3 + MNK))$, where $T$ is the upper bound of iteration. Generally, $K \ll \min(M,N)$ and $q$ is empirically small, and $h^{(c)}_{\cdot j}$s and  $w_{i\cdot}$s can be optimized in parallel. Therefore, the proposed MAP algorithm is scalable for large-scale problems.

\section{Experimental Results}
\par Here we first evaluate JMD and BJMD with two algorithms ADVI and MAP (denoted as VI-BJMD and MAP-BJMD respectively) using synthetic data. We demonstrate that both methods can estimate the variance of noise accurately and BJMD leads to superior performance compared to JMD. Then we apply our methods to three real-world data including 3-Source Text dataset, Extended Yale Face Database B and METABRIC Breast Cancer Expression dataset, and compare them with several state-of-the-art methods including $K$-means, Joint-NMF, Semi-NMF and BNMF. All experiments have been performed on a desktop computer with a 2GHz Intel Xeon E5-2683 v3 CPU, a GTX 1080 GPU card, 16GB memory and Ubuntu 16.04 (Operating System).

\subsection{Evaluation Metric}
To evaluate the effect of our methods in terms of clustering ability, we adopt an evaluation metric based on the area under the ROC curve (AUC). Specifically, $H^{(c)}$ is the coefficient matrix from source  $c$, and $L^{(c)}$ is the corresponding true label indicator matrix of $H^{(c)}$. $L^{(c)}$ is binary and $l^{(c)}_{kj}$ equals to one means that sample $x^{(c)}_{\cdot j}$ from data source $c$ belongs to the $k$th cluster. Note sample $x^{(c)}_{\cdot j}$ belongs to one or more clusters, and thus $L^{(c)}$ has no zero column. We define the following metric:
\begin{equation}
r_k^{(c)} = \max_{k=1,\ldots,K} \text{AUC}(l^{(c)}_{k\cdot}, h^{(c)}_{k\cdot})
\end{equation}
where $\text{AUC}$ takes $l^{(c)}_{k\cdot}$ as labels and $h^{(c)}_{k\cdot}$ as predictions. To evaluate the performance of model in data source $c$, we calculate $r_k^{(c)}$  from $1$ to $K$ and take average as follows:
\begin{equation} \label{metric_auc}
r^{(c)} = \frac{1}{K} \sum_{k=1}^K r_k^{(c)}
\end{equation}

\subsection{Synthetic Experiments}
\par In synthetic experiments, we compare JMD and BJMD. Note JMD is an extension of GLAD, and BJMD can be solved by two algorithms ADVI and MAP.

\subsubsection{Synthetic Data Sets}
To this end, we generate synthetic data inspired by \cite{Wu2016}. Specifically, we first generate the ground truth dictionary matrix $W_{N\times K}$ in the following step:
\begin{equation}
w_{ik} =
\begin{cases}
    a,  &  1 + (k-1)(L-coh)\leq i \leq L + (k-1)(L-coh) \quad \\
        & \text{for } k = 1 \ldots K - 1\\
    0,  & \text{otherwise}
\end{cases}
\end{equation}
where $a$ is a constant, $L$ denotes the number of non-zero entries in the first $K-1$ columns and $coh$ denotes the length of coherence between basis $w_{i-1, \cdot}$ and $w_{i\cdot}$. Note that that all entries of the $K$th column of $W$ equal to zero. Next, we generate the entries of coefficient matrix $H^{(c)}_{K\times N_c}$ for source $c$. $h^{(c)}_{kj}$ follows a Bernoulli distribution with parameter $p$, namely $h^{(c)}_{kj} \sim \text{B}(1,p)$ for $k=1,\ldots, K-1$. For the $j$th column of $H^{(c)}$, if all entries of $h^{(c)}_{\cdot j}$ equal to zero, set $h^{(c)}_{Kj} = 1$. Then we normalize $H^{(c)}$ such that the sum of column equals to one. Finally, we generate an observed data matrix by:
\begin{equation}
X^{(c)} = WH^{(c)} + \epsilon^{(c)}, \text{ where } \epsilon^{(c)} \sim \text{N}(0, \sigma_c^2)
\end{equation}

We generated synthetic data by the above procedure in both small and large scales.
\par \textbf{Samll-Scale Synthetic Data Sets:} The data were generated with $a=2$, $C=3$, $K=5$, $L=30$, $coh=5$, $N_c=120$, $p=0.3$. $\sigma_1$ and $\sigma_2$ are 1.0, 2.5 respectively and $\sigma_3$ ranges from 1.5 to 5.5 in step of 0.5. Consequently, we obtained nine datasets and each one consists of $X^{(1)}$, $X^{(2)}$, $X^{(3)}\in R^{105\times 120}$.
\par \textbf{Large-Scale Synthetic Data Sets:} Similarly, the data were generated with $a=1.5$, $C=3$, $K=10$, $L=120$, $coh=10$, $N_c=1000$, $p=0.1$. $\sigma_1$ and $\sigma_2$ are 1.0, 2.5 respectively and $\sigma_3$ ranges from 1.5 to 5.5 in step of 0.5. Consequently, we obtained nine datasets and each one  consists of $X^{(1)}$, $X^{(2)}$, $X^{(3)}\in R^{1000\times 1000}$.

\par In the following experiments, we applied JMD, VI-BJMD, MAP-BJMD to the synthetic data. We also concatenated the $X^{(1)}$, $X^{(2)}$, $X^{(3)}$ by column to get the combined data matrix. We applied GLAD, VI-BJMD and MAP-BJMD to this concatenated data matrix for comparison. We denote the results of the two algorithms as VI-catBNJMD and MAP-catBJMD respectively for clarity. The AUC value is the average of the best 5 of 20 runs (based on the values of objective function).

\subsubsection{Convergence Behavior of VI-BJMD}
\par
\begin{figure}[!t]
		\centering
		\includegraphics[width=\linewidth]{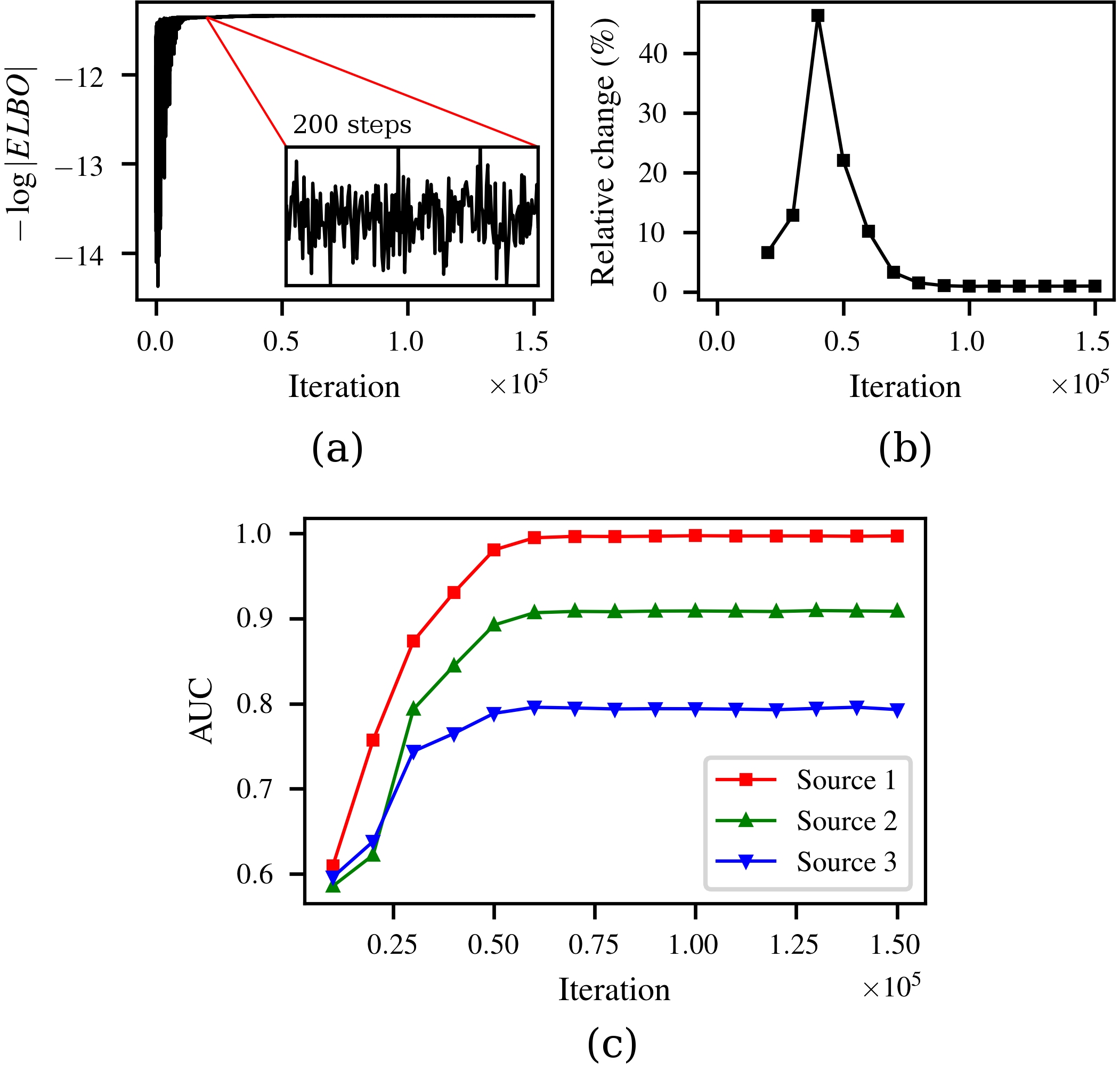}
		\caption{Illustration of the convergence behavior of VI-BJMD using a small-scale synthetic dataset with $\sigma_3$=4.0. (a) The ELBO versus iteration numbers; (b) the relative change of coefficient matrices in percentage and (c) the AUCs versus iteration numbers for each source.}
		\label{fig_vi}
\end{figure}

 We first investigate the convergence behavior of variational inference of BJMD by numerical simulation. We applied VI-BJMD onto a small-scale synthetic dataset with $\sigma_3$=4.0. We can see that the ELBO becomes stable and begins to fluctuate in a small interval after around 20,000 iterations (Fig. \ref{fig_vi}a), while the relative change of the coefficient matrices stays stable after around 80000 iterations (Fig. \ref{fig_vi}b). We also plot the steps of iteration versus the AUC performance of VI-BJMD in each source (Fig. \ref{fig_vi}c). It can be observed that the AUC performance of VI-BJMD in each source is increasing as iteration progressed. Moreover, the AUC decreases correspondingly if the noise level increases. However, the AUC performance is still increasing when the ELBO gets stable at about 20000 iterations. Therefore, if we stop iteration immediately after the ELBO becoming stable, the model has a relative poor performance in terms of AUCs. ELBO is a lower bound of evidence, which approximates the posterior distribution. Thus, it is possible that the variational distribution is approaching to the true posterior distribution, while the ELBO stays stable. As we use the coefficient matrix as predictions, it is obvious that the performance of VI-BJMD will stay stable after the relative change of coefficient matrices become stable. Therefore, our stopping criterion in (\ref{stop_advi}) of variational algorithm is a better choice than examining the change of ELBO. One should note that the relative change of coefficient matrices will not approach to zero because of the noisy gradients in optimization. Therefore, we stop iteration if the tolerance is satisfied or the maximum times of iteration $iter_{max}$ is exceeded. In the following experiments, we always set $iter_{max} = 150, 000$.

\begin{figure}[!t]
	\centering
	\includegraphics[width=\columnwidth]{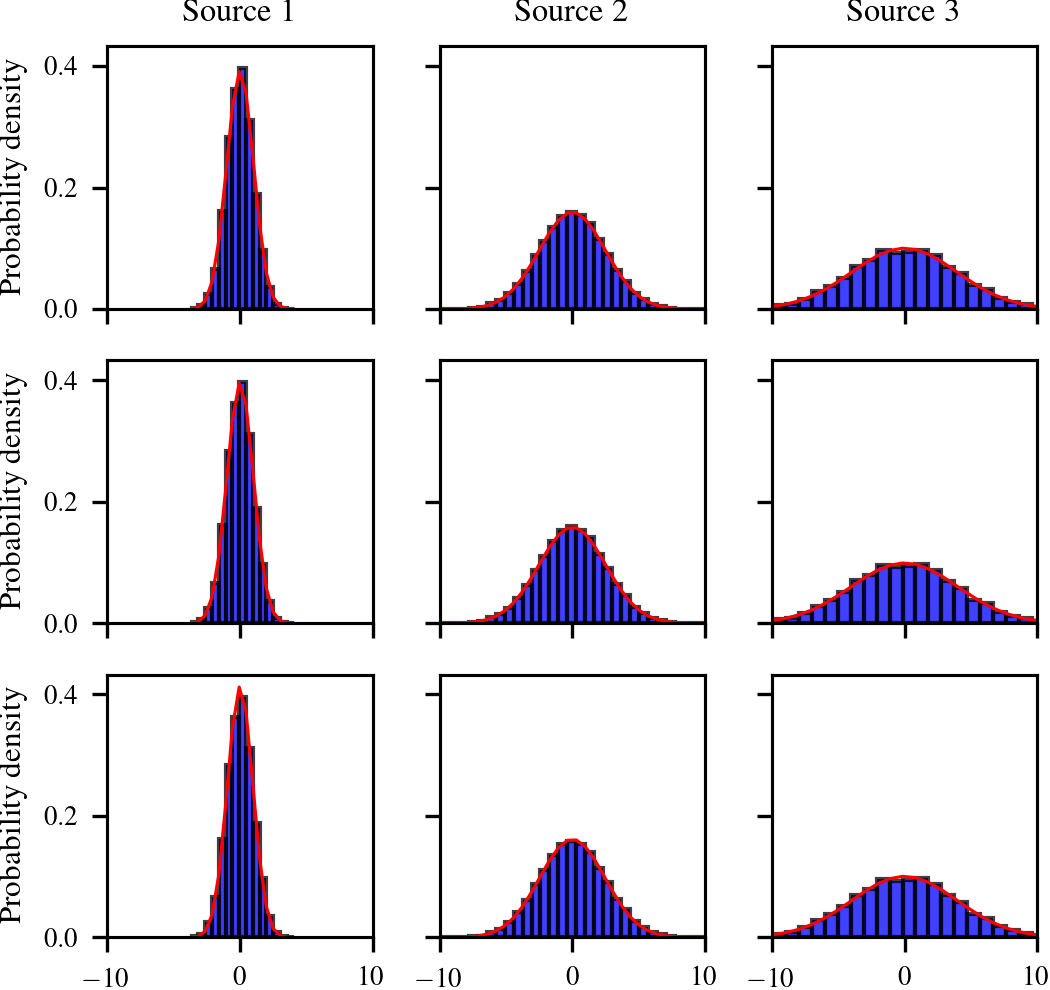}
	\caption{True noise histograms and estimated PDF (shown with curves) by JMD (top), VI-BJMD (middle) and MAP-BJMD (bottom) respectively. The true standard deviation of source 1, source 2, source 3 are 1.0, 2.5, 4.0 respectively. }\label{noise}
\end{figure}

\par Next, since our model is constructed under the heterogeneity of noise assumption, it is necessary to verify whether our proposed models can estimate the noise level of data from different sources accurately. To verify the ability of discovering the heterogeneity of noise in different data sources, we applied JMD, VI-BJMD and MAP-BJMD to the small-scale synthetic dataset with $\sigma_3$=4.0. We can clearly see that the estimated probability density curves are close to the true noise histograms, and thus all those three methods can estimate the variance of Gaussian distribution accurately (Fig. \ref{noise}).

\subsubsection{Considering the Heterogeneity of Noise Leads to Superior Performance}
\par In the following, we conduct experiments to demonstrate that considering the heterogeneity of noise in different data sources leads to superior performance. We evaluate our methods on the small and large-scale synthetic datasets both with $\sigma_3$=4.0 respectively. JMD and GLAD are omitted on the large-scale synthetic data due to their huge computational cost.

\par
\begin{figure*}[htbp]
	\centering
	\includegraphics[width=0.96\textwidth]{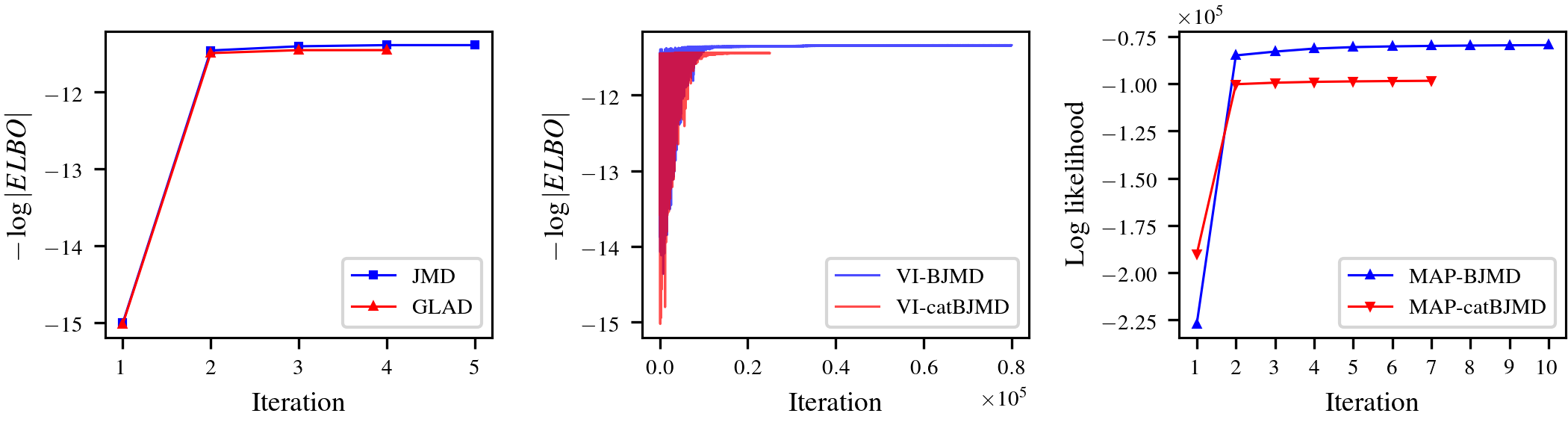}
	\caption{The objective function values versus iteration numbers for methods on a small-scale synthetic dataset with $\sigma_3$=4.0.}
	\label{fig_compare_obj_small}
\end{figure*}

 In order to facilitate comparison, we plot the objective function value ($-\log |ELBO|$, note ELBO is always negative) versus iteration number and log likelihood versus iteration number respectively (Fig. \ref{fig_compare_obj_small}). It is natural that methods considering the heterogeneity of noise get better objective function values because those methods have extra parameters to model noise of each source. Each iteration of variational inference calculates the noisy gradient by Monte Carlo sampling and can be significantly accelerated by GPU and it takes thousands of iterations to satisfy the stopping criterion. On the contrast, JMD and MAP-BJMD pay more computational cost at each iteration and converge after a few iterations.

\par We can clearly see that the methods (JMD, VI-BJMD, MAP-BJMD) considering the heterogeneity of noise demonstrate significantly better performance than the others (GLAD, VI-catBJMD, MAP-catBJMD) in terms of AUCs (TABLE \ref{table_small_scale}). Moreover, the proposed methods MAP-BJMD and VI-BJMD obtain better performance than GLAD and JMD. We can also see that both JMD and GLAD cost considerable time even in such a small synthetic data (TABLE \ref{table_small_scale}). Thus, it is impractical to apply them to large-scale real-world datasets. We utilized a GPU card to accelerate the variational inference. The synthetic data is small so that VI-BJMD can not utilize all cores of the GPU card. While concatenating all the matrices will significantly accelerate the speed of iteration. In this situation, VI-catBJMD only spends around a third of the running time of VI-BJMD. If the size of data matrix of each source is big enough, there will be no significant difference between VI-BJMD and VI-catBJMD in terms of iteration speed. MAP-BJMD and MAP-catBJMD are much faster than the other methods. Besides, MAP-BJMD gets about the same performance as VI-BJMD.

\begin{figure*}[htbp]
	\centering
	\includegraphics[width=0.66\textwidth]{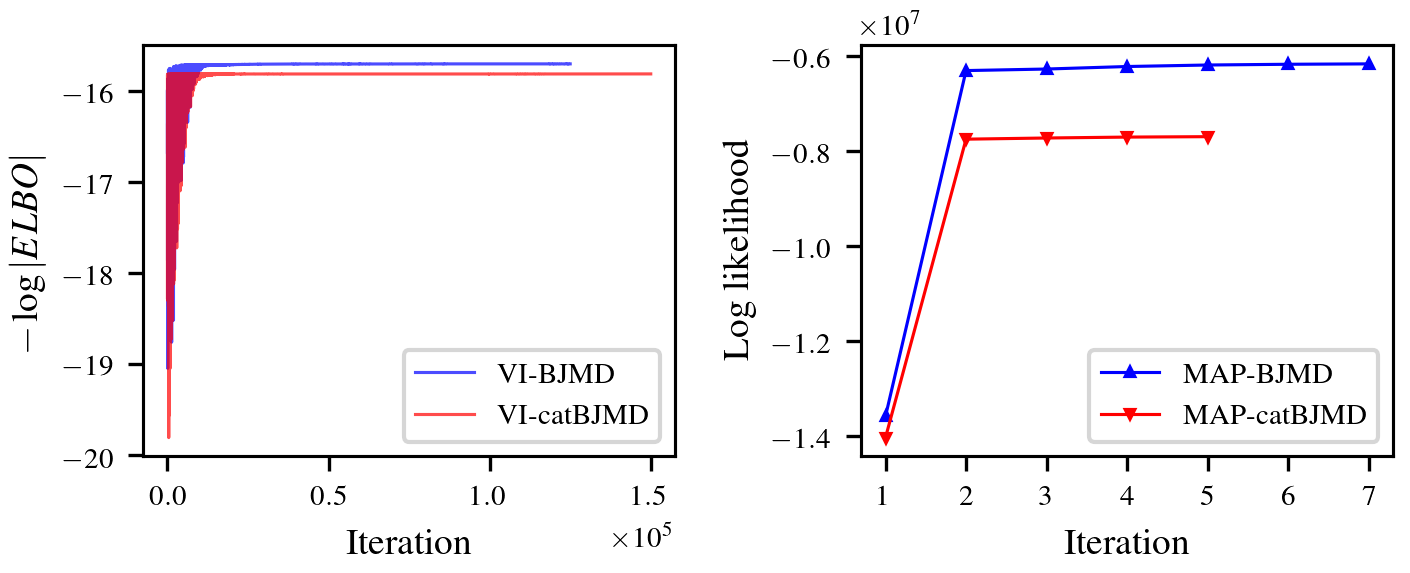}
	\caption{The objective function values versus iteration numbers for methods on a large-scale synthetic dataset with $\sigma_3$=4.0.}
	\label{fig_compare_obj_large}
\end{figure*}

\par The performance on the large-scale synthetic dataset is consistent with that on the small-scale synthetic dataset (Fig. \ref{fig_compare_obj_large} and TABLE \ref{table_large_scale}). VI-catBJMD exceeds the maximum iteration $iter_{max}$, while MAP-BJMD and MAP-catBJMD take less than ten steps to converge (Fig. \ref{fig_compare_obj_large}). Note that MAP-BJMD and VI-BJMD outperform MAP-catBJMD and VI-catBJMD respectively (TABLE \ref{table_large_scale}). Besides, VI-BJMD obtains better performance than MAP-BJMD. Since the coefficient matrices are relative large, the iteration speed of VI-catBJMD is only slightly faster than VI-BJMD. Both VI-catBJMD and VI-BJMD take more than 1000 seconds, but they are still applicable to real-world data. The MAP algorithm of BJMD are much faster than the VI one. It is more scalable for large-scale data.

\begin{table}[htbp]
	\caption{Performance Comparison on a Small-Scale Synthetic Dataset with $\sigma_3$=4.0}
	\label{table_small_scale}
	\centering
	\begin{tabular}{|c|c|c|c|c|}
		\hline
		 \multirow{2}{*}{Algorithm} & \multicolumn{3}{c|}{AUC}  & \multirow{2}{*}{Time (s)} \\  \cline{2-4}
             		                & Source 1 & Source 2 & Source 3  &        \\
		\hline	
		GLAD & 85.25 & 74.32 & 66.35 & 5068.10\\
		\hline
		JMD & 90.23 & 80.96 & 72.06 & 5709.29\\
		\hline\hline
		VI-catBJMD & 86.99 & 78.51 & 72.79 & 84.44 \\
		\hline
		VI-BJMD & 99.66 & 90.87 & 79.16 & 230.21\\
		\hline\hline
		MAP-catBJMD & 92.04 & 82.98 & 69.79 & 5.81\\
		\hline
		MAP-BJMD & 99.67 & 90.31 & 77.85 & 4.20 \\
		\hline		
	\end{tabular}
\end{table}

\begin{table}[htbp]
	\caption{Performance Comparison on a Large-Scale Synthetic Dataset with $\sigma_3$=4.0}
	\label{table_large_scale}
	\centering
	\begin{tabular}{|c|c|c|c|c|}
		\hline
		\multirow{2}{*}{Algorithm} & \multicolumn{3}{c|}{AUC}  & \multirow{2}{*}{Time (s)} \\  \cline{2-4}
		& Source 1 & Source 2 & Source 3  &        \\
		\hline
		VI-catBJMD & 89.46 & 84.26 & 77.97 & 1461.89 \\
		\hline
		VI-BJMD & 99.84 & 96.96 & 91.85 & 1154.06\\
		\hline\hline
		MAP-catBJMD & 89.79 & 80.16 & 72.45 & 27.26\\
		\hline
		MAP-BJMD & 98.20 & 94.29 & 89.09 & 43.16 \\
		\hline		
	\end{tabular}
\end{table}

\subsubsection{The Effects of Heterogeneous Noise}
\par
\begin{figure*}[htbp]
	\centering
	\includegraphics[width=0.9\textwidth]{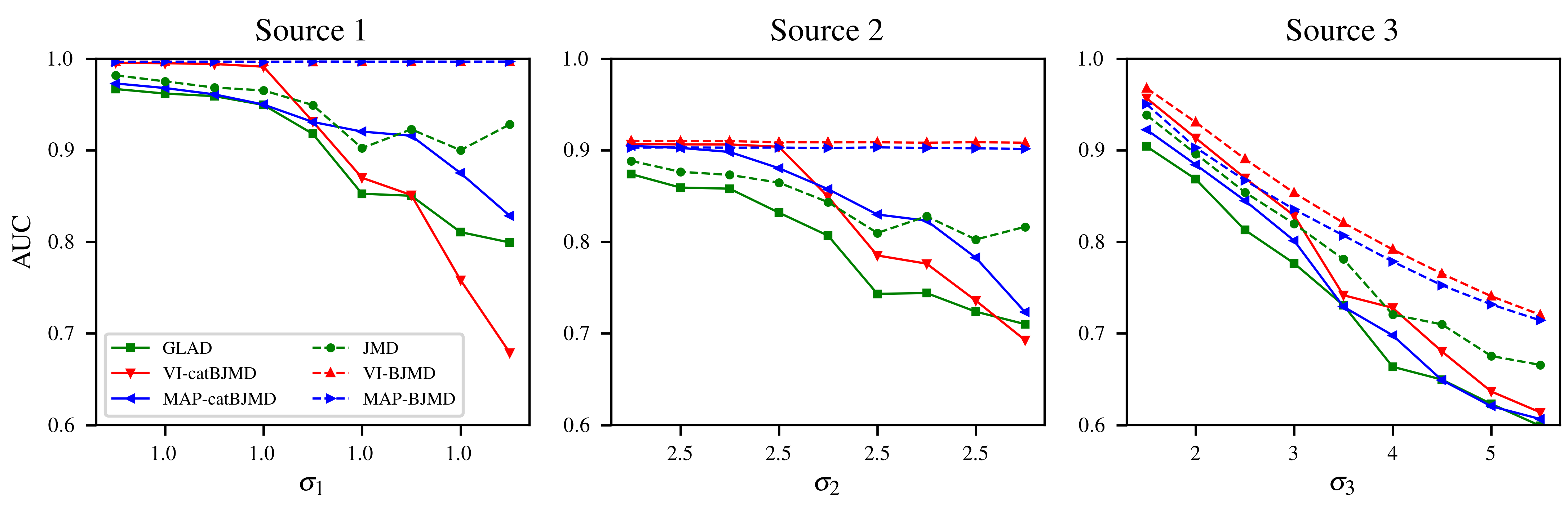}
	\caption{Performance comparison of JMD, VI-BJMD and MAP-BJMD and their respective version ignoring heterogeneous noise on the small-scale synthetic datasets. The standard deviation of Gaussian noise is fixed in Source 1 and Source 2. The noise level of Source 3 ranges from 1.5 to 5.5. Performance of algorithms (JMD, VI-BJMD, MAP-BJMD) considering difference of noise level is plotted in dashed line and the others are plotted in solid line. }
	\label{fig_small_auc}
\end{figure*}

 To further investigate the effects of heterogeneous noise on model performance, we evaluate our methods on the small and large-scale synthetic datasets respectively. JMD and GLAD are omitted on the large-scale synthetic datasets. The results are showed in Fig \ref{fig_small_auc}. We can clearly see that:
\begin{itemize}
\item Considering the heterogeneity of noise leads to superior performance. For example, AUCs of JMD are always higher than those of GLAD, and it is the same for VI-BJMD and MAP-BJMD.
\item The gap of performance between the algorithm considering heterogeneity of noise and those ignoring it is widened with the increase of noise level in Source 3.
\item AUCs of VI-BJMD and MAP-BJMD in Source 1 and Source 2 are not influenced by the increasing noise level in Source 3, which indicates that our proposed algorithms are more suitable for integrating data with different noise level.
\item The AUCs of VI-catBJMD in Source 1 and Source 2 suffers rapid decreasing after the noise level of Source 3 is greater than 2.5. The reason is that when the noise level of Source 3 is greater than 2.5, the variances of noise in Source 1 and Source 2 are overestimated, and thus the gradients of the coefficient matrices of Source 1 and Source 2 are smaller than the true gradient. As a consequence, VI-catBJMD needs much more iterations. 
\item AUCs of VI-BJMD are slightly better than those of MAP-BJMD. VI-BJMD approximates the posterior distribution by variational distribution, while MAP-BJMD approximates the posterior distribution by its mode of density function. As a result, VI should more robust than MAP when the model is misspecified. One can observe that the AUCs of VI-catBJMD are better than those of MAP-catBJMD, when the noise level of Source 3 is less than 3.0.
\end{itemize}

\par
\begin{figure*}[htbp]
	\centering
	\includegraphics[width=0.9\textwidth]{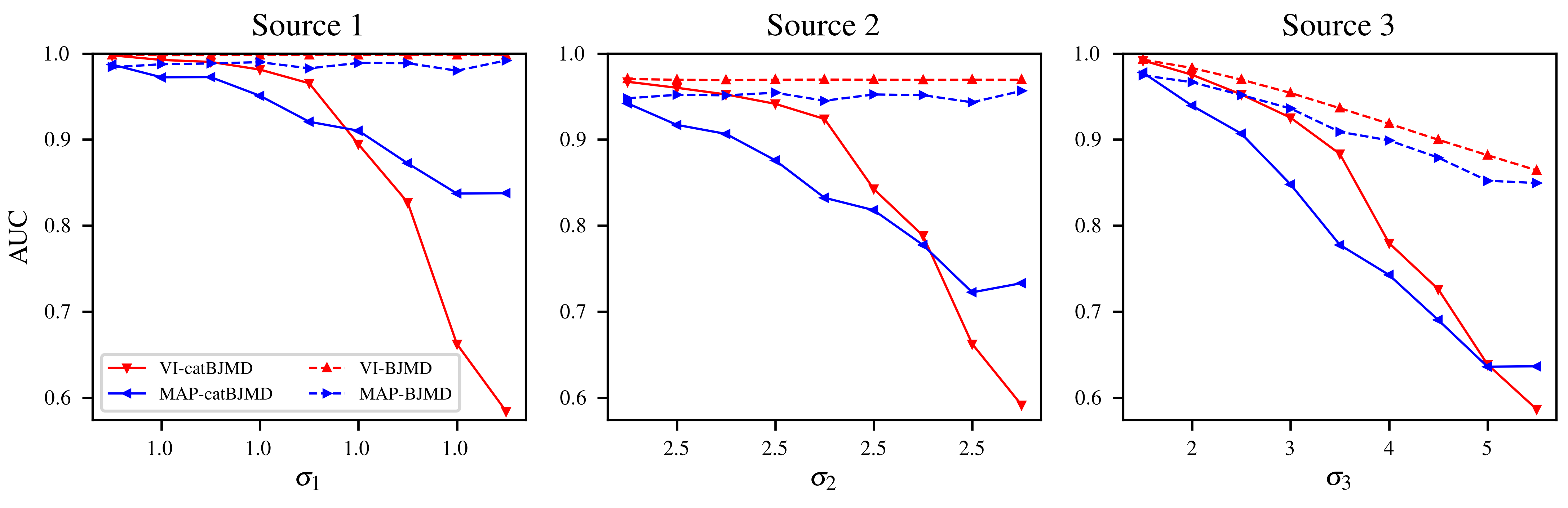}
	\caption{Performance comparison of VI-BJMD and MAP-BJMD (as well as VI-catBJMD, MAP-catBJMD) on the large-scale synthetic datasets. The standard deviation of Gaussian noise is fixed in Source 1 and Source 2. The noise level of Source 3 ranges from 1.5 to 5.5.}
	\label{fig_large_auc}
\end{figure*}

 The results on the large-scale synthetic datasets are consistent with those on the small-scale ones (Fig \ref{fig_large_auc}). We can still observe that considering the heterogeneity of noise leads to superior performance. When the noise level on Source 3 is more than 3.5, the AUC performance of VI-catBJMD decreases sharply. Compared to the result on small-scale ones, VI-BJMD outperforms MAP-BJMD significantly. The performance curves of VI-BJMD on Source 1 and Source 2 are very stable while the noise level of Source 3 increasing. It implies that VI-BJMD successfully protects Source 1 and Source 2 from the influence of highly noisy source. However, the performance curves of MAP-BJMD fluctuates slightly. Moreover, the performance of VI-BJMD is consistently better than MAP-BJMD in all sources.

\begin{table*}[hpbt]
	\caption{Summary of the Three Data Sets}
	\label{table_stats}
	\centering
	\begin{tabular}{|c||c|c|c|c|}
		\hline
		Dataset	& Dimensionality ($M$)	&  Size ($N$) & $\#$ of groups & $\#$ of classes\\		
		\hline
		3Sources & 4750 &  948 & 3 & 6	\\			
		\hline
		Extended Yale Face Database B &  2016&  1244 & 2 & 38	\\
		\hline
		METABRIC & 19466 &   1911& 3 & 5	\\
		\hline		
	\end{tabular}
\end{table*}

\subsection{Real-World Experiments}
\par In this subsection, we applied VI-BJMD and MAP-BJMD to three real-world datasets and compared their performance with $K$-means, joint-NMF, BNMF and semi-NMF. GLAD and JMD are omitted due to their huge computational cost. The three datasets and data preprocessing are summarized below (TABLE \ref{table_stats}).

\par \textbf{3Sources Dataset}: The first dataset is the so-called 3Sources text data obtained from \url{http://mlg.ucd.ie/datasets/3sources.html}. This dataset is collected from three well-known online news sources: BBC, Reuters and The Guardian, which contains 948 news articles covering 416 distinct news stories. Each story is annotated manually with one or more of six labels. BBC, Reuters and The Guardian are represented by term-document matrices of size $3560\times 352$, $3068\times 294$, $3631\times 302$ respectively. To conduct an integrative clustering, we combined the terms of those data matrices. As a result, the data matrices are of sizes $4750\times 352$, $4750\times 294$, $4750 \times 302$ respectively. Note that this dataset is very sparse. Each article is represented by a 4750-dimensional vector and only a few entries are non-zeros.

\par \textbf{Extended Yale Face Database B Dataset}: The next dataset is the so-called Extended Yale Face Database B data \cite{georghiades2001few, lee2005acquiring}, which contains images of 38 individuals collected under different illumination conditions. For each image, the illumination condition is recorded by azimuth and elevation angles of the source of light. We used images whose azimuth and elevation angles are smaller than 60 degrees and resized them to $42\times 48$. Those images can be divided into two groups: center lighting, the source of light is near to the center with azimuth angle $< 20^\circ$ and there are $567$ images in total; side lighting, the light is from side with azimuth angle $> 20^\circ$ and there are $677$ images in total. Therefore, the data matrices for center lighting and side lighting are of sizes $2016\times 567$ and $2016\times 677$ respectively. We normalized the data matrices by dividing its average for further analysis. Images collected under side lighting have more shadow, and thus it is natural to assumes that the noise level of the two groups is different.
\begin{table*}[hpbt]
	\caption{The Noise Level Estimated in Each Dataset}
	\label{table_noise_level}
	\centering
	\begin{tabular}{|c||c|c|c||c|c||c|c|c||c|c|c|}
		\hline
		\multirow{2}{*}{Algorithm} & \multicolumn{3}{c||}{3Sources} & \multicolumn{2}{c||}{Yale Face B} & \multicolumn{3}{c||}{METABRIC} &  \multicolumn{3}{c|}{Filtered METABRIC} \\
		\cline{2-12}
		& BBC	& Guardian & Reuters & Center & Side  & Low	& Moderate & High & Low	& Moderate & High \\
		\hline
		VI-BJMD  & 0.3226 & 0.3390 &  0.2922 & 0.1186& 0.1804&-       &-       & - &   0.5987 & 0.5998 & 0.6348 \\
		\hline
		MAP-BJMD &0.3210 & 0.3370  &  0.2909 & 0.1106& 0.1790& 0.3455 & 0.3454 & 0.3684& 0.5987& 0.5987 &0.6346\\
		\hline
	\end{tabular}
\end{table*}

\begin{table*}[hpbt]
	\captionsetup{justification=centering}
	\begin{minipage}{.5\linewidth}
		\caption{Performance Comparison of Different Methods \\ on 3Sources Dataset}
		\label{table_3source_performance}
		\centering
		\begin{tabular}{|c|c|c|c|c|}
			\hline
			\multirow{2}{*}{Algorithm} & \multicolumn{3}{c|}{AUC (\%)}  & \multirow{2}{*}{Time (s)} \\  \cline{2-4}
			& BBC	& Guardian & Reuters &        \\
			\hline
			$K$-means & 65.60 & 66.14 & 65.60 & 4.40 \\
			\hline
			Joint-NMF & 80.95 & 80.61 & 80.94 & 0.65\\
			\hline
			Semi-NMF & 76.42 & 73.70 & 76.42 & 41.99\\
			\hline
			BNMF & 82.84 & 83.51 & 82.87 & 77.37\\
			\hline
			VI-BJMD & 90.25 & 90.78 & 90.33 & 2274.80\\
			\hline
			MAP-BJMD & 90.57 & 88.82 & 89.46 & 6.55\\
			\hline		
		\end{tabular}
	\end{minipage}%
	\begin{minipage}{.5\linewidth}
		\caption{Performance Comparison of Different Methods \\ on Extended Yale Face B Dataset}
		\label{table_face_performance}
		\centering
		\begin{tabular}{|c|c|c|c|c|}
			\hline
			\multirow{2}{*}{Algorithm} & \multicolumn{2}{c|}{AUC (\%)}  & \multirow{2}{*}{Time (s)} \\  \cline{2-3}
			& Center & Side  &        \\
			\hline
			$K$-means & 61.97 & 55.77  & 6.05  \\
			\hline
			Joint-NMF & 93.69 & 83.37   & 3.91 \\
			\hline
			Semi-NMF & 96.81 & 91.98    & 53.79\\
			\hline
			BNMF & 94.63 & 85.50  &  189.55 \\
			\hline
			VI-BJMD & 97.11 & 90.56  & 1444.34\\
			\hline
			MAP-BJMD & 97.47  & 92.58  &  112.20\\
			\hline		
		\end{tabular}
	\end{minipage}
\end{table*}

\par \textbf{METABRIC Breast Cancer Dataset}: The third dataset is the so-called METABRIC data \cite{curtis2012genomic, liu2014breast}, which contains a large breast cancer patient cohort of around 2000 samples with detailed clinical measurements and genome-wide molecular profiles. Using gene expression to classify invasive breast cancers into biologically and clinically distinct subtypes has been studied for over a decade. In 2009, Parker \textit{et al}. developed an efficient classifier called PAM50, to distinguish the intrinsic subtypes of breast cancer \cite{parker2009supervised}. PAM50 is now commonly employed in breast cancer studies \cite{bastien2012pam50, nielsen2010comparison, ellis2011randomized}. Thus, we used the PAM50 subtypes as reference to evaluate the results of clustering. The cellularity of breast tumor is a concerned clinical indicator in treatment \cite{chang2005patterns, rajan2004change, hatakenaka2008apparent}. Therefore, we used the cellularity to divide samples into groups. We first mapped probes of gene expression to gene names by average pooling. Then we removed samples that PAM50 subtypes or cellularity is not available. There are 19466 genes and 1911 samples left after preprocessing. The samples are divided into 3 groups: low cellularity, moderate cellularity and high cellularity with size of $214$, $735$ and $962$, respectively. The number of features of this dataset is much larger than the number of samples.

\par  We applied our algorithms to these three datasets. In all experiments, we set the Dirichlet prior $\alpha_{0k}=1.1$ and the Laplace prior $\lambda=1.0$. The tolerance of MAP-BJMD and VI-BJMD is $10^{-3}$ and $10^{-2}$, respectively. We used $K$-means, joint-NMF, BNMF and semi-NMF for comparison. Among these four algorithms, BNMF is a Bayesian approach. For each experiment, the AUC is reported by the average of the best 5 of 20 runs.

\par TABLE \ref{table_noise_level} shows the estimated noise level on each dataset. TABLE \ref{table_3source_performance}, \ref{table_face_performance} and \ref{table_metabric_performance} show the clustering performance on 3Sources, Extended Yale Face Database B and METABRIC dataset, respectively. The AUCs and running time are reported. VI-BJMD takes much time on the METABRIC dataset and thus it is not reported (TABLE \ref{table_metabric_performance}).
\par These experiments reveal some important and interesting points:
\begin{itemize}
	\item Both MAP-BJMD and VI-BJMD can achieve similar or better performance than the competing methods over all datasets.
	\item The algorithm of Semi-NMF involves calculating the inverse of matrix. As we mentioned earlier, 3Sources data is very sparse. Therefore, Semi-NMF is not robust on this dataset. MAP-BJMD involves calculating the inverse of matrix in updating basis matrix too, but the Laplace priors can protect the inverse of matrix from singular. The Bayesian methods including BNMF, VI-BJMD and MAP-BJMD outperform other methods on 3Sources dataset. It suggests the superiority of Bayesian priors for very sparse data.
	\item All methods except MAP-BJMD have relative poor performance on METABRIC data. This suggests that MAP-BJMD is more robust than other methods. We note that the number of features is much larger than the number of samples in the METABRIC dataset. We believe that feature selection will be needed to improve this situation.
	\item  Although VI-BJMD outperforms MAP-BJMD in synthetic experiments. However, VI-BJMD is not guaranteed to outperform MAP-BJMD in real-wold data. Real-world data is much more complicated and VI-BJMD often exceeds the maximum steps.
	\item  Bayesian methods including VI-BJMD and BNMF take more time than other methods, while the speed of MAP-BJMD is comparable to other ones. Thus, MAP-BJMD is more scalable and efficient.
	\item  The estimated noise levels of VI-BJMD and MAP-BJMD are very consistent. Moreover, the estimated noise level can be used to guide feature selection as discussed in the next subsection.
\end{itemize}

\subsection{Adaptive Feature Selection}
\par In this subsection, we discuss how to use the estimated noise level to select features adaptively. In biological applications, the features are often redundant. It is common that we have thousands of features but only tens or hundreds of samples. Thus, feature selection is needed. Perhaps the most straightforward approach to select features is to choose those variance greater than a threshold. However, it is difficult to specify the threshold value for unsupervised tasks. Luckily, we can use the estimated noise level as the threshold to conduct an adaptive feature selection.

\begin{table*}[hpbt]
	\captionsetup{justification=centering}
	\begin{minipage}[t]{.5\linewidth}
		\caption{Performance Comparison of Different Methods \\ on METABRIC Dataset}
		\label{table_metabric_performance}
		\centering
		\begin{tabular}{|c|c|c|c|c|}
			\hline
			\multirow{2}{*}{Algorithm} & \multicolumn{3}{c|}{AUC (\%)}  & \multirow{2}{*}{Time (s)} \\  \cline{2-4}
			& Low	& Moderate & High &        \\
			\hline
			$K$-means & 62.99 & 66.82 & 72.73& 24.74 \\
			\hline
			Joint-NMF & 65.81 & 71.85 & 75.07 & 45.58\\
			\hline
			Semi-NMF & 58.90 & 55.02 & 53.48 & 311.24\\
			\hline
			BNMF & 64.56 & 66.27 & 68.59 & 423.15\\
			\hline
			MAP-BJMD & 82.22 & 79.05 & 81.45  &  27.17\\
			\hline
		\end{tabular}
	\end{minipage}%
  \begin{minipage}[t]{.5\linewidth}
		\caption{Performance Comparison of Different Methods \\ on Filtered METABRIC Dataset}
		\label{table_fmetabric_performance}
		\centering
		\begin{tabular}{|c|c|c|c|c|}
			\hline
			\multirow{2}{*}{Algorithm} & \multicolumn{3}{c|}{AUC (\%)}  & \multirow{2}{*}{Time (s)} \\  \cline{2-4}
			& Low	& Moderate & High &        \\
			\hline
			$K$-means & 67.01 & 70.16 & 78.85&  5.53 \\
			\hline
			Joint-NMF & 82.70 & 79.93 & 82.51 & 18.76\\
			\hline
			Semi-NMF & 63.18 & 62.89 & 59.89 & 76.28\\
			\hline
			BNMF & 82.71 & 80.32 & 85.43 & 101.96\\
			\hline
			VI-BJMD & 84.93 & 81.81 &85.16& 3623.24\\
			\hline
			MAP-BJMD & 84.34& 79.55 & 84.03&20.70\\
			\hline
		\end{tabular}
	\end{minipage}
\end{table*}

\par Specifically, as MAP-BJMD is much faster than VI-BJMD, we can use MAP-BJMD to estimate the variance of Gaussian noise for each source. If the variance of a feature is less than the variance of noise. It is natural to assume that it has no significant effect on clustering. The procedure of adaptive feature selection is summarized as follows: Given a dataset with C sources, we first use MAP-BJMD to estimate the variance $\sigma_c^2$ of each source. Then, for each source we conduct $M_c$ individual hypothesis tests with null hypothesis $H_0$: variance of feature $p$ $\sigma_{cp}^2 = \sigma_c^2$ and one tailed  alternative  hypothesis $\sigma_{cp}^2 > \sigma_c^2$. Finally, given significant level, we simply use the Bonferroni correction  to choose the features that variances are significant greater than background noise among all sources. We applied this procedure to METABRIC dataset. There are 4210 remaining genes out of 19466 after this selection with adjusted $q$-value$<0.05$. 

\par As shown in TABLE \ref{table_fmetabric_performance}, the feature selection improves the performance of all methods compared with TABLE \ref{table_metabric_performance}. It implies that our feature selection procedure is effective. VI-BJMD and MAP-BJMD outperform other algorithms. Moreover, the feature selection procedure reduces the size of data matrices. Therefore, the running time of VI-BJMD is acceptable and it outperforms all other methods in terms of AUCs.

\section{Discussion and Conclusion}
\par In this paper, we have proposed a Bayesian joint matrix decomposition framework, which models the heterogeneity of noise explicitly by Gaussian distribution. We develop two algorithms VI and MAP to solve this framework. VI approximates the posterior more accurately, while MAP is more scalable. Both algorithms can estimate the noise level of each data source accurately. Experimental results on synthetic datasets show that considering the heterogeneity of noise of different sources brings an improvement in clustering performance and protects the data source of low noise level from influence due to noisy data sources. Experiments on real-world datasets demonstrate that our methods can achieve better or competitive performance than the state-of-the-art methods.

\par It is well to be reminded that several questions remain to be investigated in future studies. First, VI-BJMD consistently outperforms the MAP-BJMD in synthetic experiments, but it is not distinctly observed on the real-world data experiments. It possibly implies that ADVI does not work well on the real data, which is more complicated than the synthetic data. It remains a problem how to infer such a model in a more efficient and effective way by variational methods. Second, BJMD models the heterogeneous noise by Gaussian distribution. However, the noise in real-world might be structured and more complicated. Instead of assuming the noise distribution follows a family of parametric distribution, nonparametric Bayesian methods are more flexible and might fit more to the complicated noise in the real-world data. This may suggest a way to extend the current BJMD model.


%

\appendices
\ifCLASSOPTIONcompsoc
  \section*{Acknowledgments}
\else
  \section*{Acknowledgment}
\fi

This work has been supported by the National Natural Science Foundation of China [No. 61422309, 61379092, 61621003 and 11661141019]; the Strategic Priority Research Program of the Chinese Academy of Sciences (CAS) [XDB13040600] and CAS Frontier Science Research Key Project for Top Young Scientist [No. QYZDB-SSW-SYS008].

\ifCLASSOPTIONcaptionsoff
  \newpage
\fi



%
\bibliographystyle{IEEEtran}
\bibliography{IEEEabrv,reference}

\end{document}